\documentclass{article} 
\usepackage{iclr2024_conference,times}

\usepackage[utf8]{inputenc} 
\usepackage[T1]{fontenc}    
\usepackage{hyperref}       
\usepackage{url}            
\usepackage{booktabs}       
\usepackage{amsfonts}       
\usepackage{nicefrac}       
\usepackage{microtype}      
\usepackage{titletoc}

\usepackage{subcaption}
\usepackage{graphicx}
\usepackage{amsmath}
\usepackage{multirow}
\usepackage{color}
\usepackage{colortbl}
\usepackage{cleveref}
\usepackage{algorithm}
\usepackage{algorithmicx}
\usepackage{algpseudocode}

\graphicspath{{../}} 

\title{Time-Aware Feature Selection: Adaptive Temporal Masking for Stable Sparse Autoencoder Training}

\author{%
T. Ed Li$^{*1}$, Junyu Ren$^{*2}$\\[1ex]
Yale University$^{1}$, University of Chicago$^{2}$\\[1ex]
\textsuperscript{*}Equal contribution
}

\begin{document}

\maketitle

\begin{abstract}
Understanding the internal representations of large language models is crucial for ensuring their reliability and safety, with sparse autoencoders (SAEs) emerging as a promising interpretability approach. However, current SAE training methods face feature absorption, where features (or neurons) are absorbed into each other to minimize $L_1$ penalty, making it difficult to consistently identify and analyze model behaviors. We introduce Adaptive Temporal Masking (ATM), a novel training approach that dynamically adjusts feature selection by tracking activation magnitudes, frequencies, and reconstruction contributions to compute importance scores that evolve over time. ATM applies a probabilistic masking mechanism based on statistical thresholding of these importance scores, creating a more natural feature selection process. Through extensive experiments on the Gemma-2-2b model, we demonstrate that ATM achieves substantially lower absorption scores compared to existing methods like TopK and JumpReLU SAEs, while maintaining excellent reconstruction quality. These results establish ATM as a principled solution for learning stable, interpretable features in neural networks, providing a foundation for more reliable model analysis.
\end{abstract}

\section{Introduction}
\label{sec:intro}

Large language models (LLMs) have demonstrated remarkable capabilities across a wide range of tasks, yet understanding their internal representations remains crucial for ensuring their reliability and safety \citep{gpt4}. Sparse autoencoders (SAEs) offer a promising approach by decomposing neural activations into interpretable features \citep{cunningham2023}, but face a fundamental challenge: existing methods rely on rigid sparsity constraints that often lead to unstable or non-interpretable features.

\textit{Feature Absorption} \citep{chaninAbsorptionStudyingFeature2024} is a phenomenon observed in Sparse Autoencoders (SAEs) where one latent dimension effectively subsumes multiple distinct features that share an implication relationship, thereby enhancing sparsity at the cost of interpretability. A typical example arises when every token labeled ``short'' also starts with ``S'': in this case, the SAE can discard the separate ``starts with S'' feature by embedding it into the ``short'' latent. Similar behavior emerges in hierarchical taxonomies (e.g.,``India'' implying ``Asia''), since a broader concept can encompass the subordinate feature in a single latent dimension. However, such merging can lead to inconsistent or partial coverage of features: some tokens or classes expected to activate in a particular dimension might be omitted.

We introduce Adaptive Temporal Masking (ATM), which rethinks feature selection through temporal dynamics. Our approach tracks activation magnitudes, frequencies, and reconstruction contributions through exponential moving averages to compute importance scores that evolve over training. We apply a probabilistic masking mechanism based on statistical thresholding of these importance scores, creating a more natural feature selection process than hard thresholds. This allows ATM to automatically adapt to activation patterns while maintaining feature stability.

Our main contribution, ATM, is a combination of:
\begin{itemize}
    \item A novel temporal masking mechanism that significantly reduces feature absorption while maintaining strong reconstruction quality.
    \item A principled importance-based feature selection approach that combines activation magnitudes and reconstruction contributions.
    \item A probabilistic masking scheme based on statistical thresholding that adapts to evolving feature importance distributions.
    \item A comprehensive evaluation showing ATM's effectiveness across multiple metrics.
\end{itemize}

The implications of this work extend beyond technical improvements in SAE training. By providing more stable and interpretable features, ATM enables more reliable analysis of LLM internals, crucial for understanding model behavior and potential biases \citep{de-arteagaBiasBiosCase2019}.

\definecolor{purpleRGB}{RGB}{127,0,127} 

\begin{figure}[h]
\begin{center}
    \includegraphics[width=\textwidth]{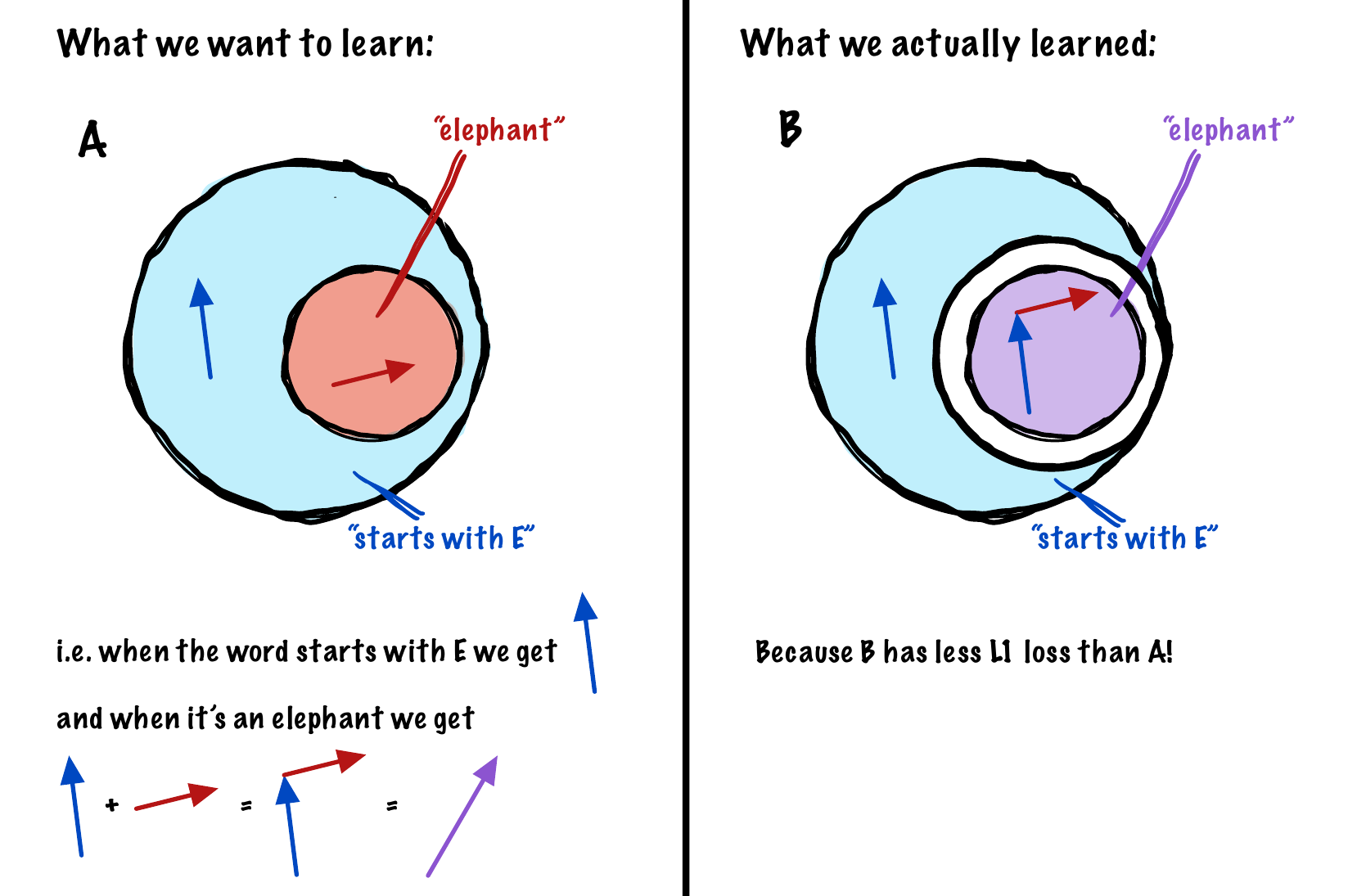} 
\end{center}
\caption{\textbf{Visualization of feature absorption.} Panel A represents the target scenario, where the SAE learns two features in two neurons: \textcolor{blue}{``starts with E'' (blue)} and \textcolor{red}{``elephant'' (red)}. When the underlying token is <elephant>, both neurons should light up resulting in an overall purple activation vector for token \color{purpleRGB}{<elephant>} \textcolor{black}{. However, panel B reveals what the SAE actually learns due to $L_1$ loss: the ``elephant'' feature can absorb the ``starts with E'' feature, which effectively reduces the number of active latents (lower $L_1$ norm) when the underlying token is \textit{<elephant>}. While this increases sparsity, it diminishes interpretability since the ``starts with E'' feature no longer activates independently. Instead, the ``elephant'' feature acquires an unintended downstream effect, making feature activations less modular. The figure is adapted from \href{https://colab.research.google.com/drive/1ePkM8oBHIEZ2kcqAiA3waeAmz8RSdHmq}{ARENA Tutorials} \citet{colab2025}.}}

\label{fig:pdf-example}
\end{figure}



\section{Background}
\label{sec:background}

Deep learning models, particularly large language models (LLMs), have demonstrated remarkable capabilities but remain challenging to interpret \citep{goodfellow2016deep,Ribeiro2016WhySI,Doshi-Velez2017TowardsAR}. While various methods have been developed to understand their internal representations \citep{Montavon2017MethodsFI}, sparse autoencoders (SAEs) have emerged as a particularly promising approach by decomposing neural activations into interpretable features \citep{gaoScalingEvaluatingSparse,Le2011BuildingHF}.

The development of SAE training methods has focused on balancing sparsity with feature interpretability. Early approaches used simple $\ell_1$ regularization, while recent methods like TopK SAEs \citep{gaoScalingEvaluatingSparse} employ hard thresholding. JumpReLU \citep{rajamanoharanJumpingAheadImproving2024} introduced discontinuous activation functions to improve stability, and Switch SAEs \citep{mudideEfficientDictionaryLearning2024a} explored learned routing mechanisms. However, these architectural solutions often struggled with feature absorption \citep{chaninAbsorptionStudyingFeature2024}, where distinct concepts become entangled.

\subsection{Problem Setting}
\label{subsec:problem}

Feature absorption is quantified following the methodology of \citet{chaninAbsorptionStudyingFeature2024}, which relies on a first-letter classification task. Specifically, tokens (composed of English letters and optional leading spaces) are split into training and test sets, and a supervised logistic regression probe is trained on the model's latent activations. Subsequently, $k$-sparse probing is used to identify the latents relevant to the classification task. A feature is designated as ``absorbed'' if the principal feature-split latents, as determined by the threshold $\tau_{\mathrm{fs}} = 0.03$, fail to classify the feature, yet the logistic regression probe still succeeds by activating a different latent that: (i) exhibits a cosine similarity of at least $\tau_{\mathrm{ps}} = 0.025$, and (ii) explains at least $\tau_{\mathrm{pa}} = 0.4$ of the probe's projection. \par
Notably, this approach utilizes probe projection analyses instead of latent ablation, thereby allowing the identification of absorbed features across all layers of the SAE --- including those where the relevant information has migrated from its original location to later stages. As a result, this criterion provides a broader and more robust benchmark for detecting subtle absorption behaviors that might remain undetected by simpler methods. \par
The key difficulty lies in balancing feature stability with sparsity requirements. Past SAEs \citep{cunningham2023,gaoScalingEvaluatingSparse,rajamanoharanJumpingAheadImproving2024} achieve sparsity through hard thresholding but suffer from high feature absorption, where semantically distinct concepts become entangled \citep{chaninAbsorptionStudyingFeature2024}.\par

Given activation vectors $\mathbf{x} \in \mathbb{R}^d$ from a transformer layer with hidden dimension $d$, we aim to learn an encoder $E: \mathbb{R}^d \rightarrow \mathbb{R}^n$ and decoder $D: \mathbb{R}^n \rightarrow \mathbb{R}^d$ that minimize reconstruction error while maintaining sparsity:

\begin{align}
    \min_{E,D} \mathbb{E}_{\mathbf{x}}\left[\|D(E(\mathbf{x})) - \mathbf{x}\|_2^2\right] \quad \text{subject to } \|E(\mathbf{x})\|_0 \ll n \label{eq:objective}
\end{align}

The encoder maps inputs to a higher-dimensional space ($n > d$) to allow for overcomplete feature learning, while the decoder reconstructs the original activations. The $\ell_0$ constraint in Equation~\ref{eq:objective} ensures sparsity.

Our approach makes two key assumptions about neural activation patterns:
\begin{enumerate}
    \item \textbf{Temporal Consistency}: Feature importance follows predictable patterns over time, enabling reliable tracking through exponential moving averages
    \item \textbf{Statistical Structure}: Activation distributions exhibit regularities that can be leveraged for adaptive thresholding
\end{enumerate}

These assumptions motivate our temporal masking mechanism, which differs from previous work by focusing on the dynamic nature of feature importance rather than architectural modifications. This perspective enables more principled feature selection while maintaining the computational efficiency of standard SAEs.

\section{Method}
\label{sec:method}

Building on the formalism introduced in Section~\ref{subsec:problem}, we propose Adaptive Temporal Masking (ATM) to solve the sparse autoencoding problem in Equation~\ref{eq:objective}. Our approach replaces rigid sparsity constraints with dynamic feature selection based on temporal activation patterns, addressing both assumptions identified in Section~\ref{subsec:problem}: temporal consistency and statistical structure.

\subsection{Temporal Importance Score Tracking}
\label{subsec:importance}
To capture the evolving significance of features during training, we track three key statistics using exponential moving averages (EMAs):
\begin{align}
    \text{Magnitude EMA}(t) &= \beta \cdot \text{Magnitude EMA}(t-1) + (1-\beta) \cdot \mathbb{E}_\text{batch}[|f_t|] \label{eq:magnitude_ema} \\
    \text{Recon. Contrib.}(t) &= \beta \cdot \text{Recon. Contrib.}(t-1) + (1-\beta) \cdot \left|\frac{\partial \mathcal{L}_\text{recon}}{\partial f_t}\right| \label{eq:recon_ema}
\end{align}
where $\beta = 0.9$ is the decay rate, $f_t$ represents feature activations at time $t$, and $\mathbb{E}_\text{batch}$ denotes the expectation over the current batch. The reconstruction contribution term measures the gradient magnitude of the reconstruction loss with respect to each feature, capturing how much each feature contributes to minimizing the reconstruction error. These statistics are combined to form an importance score:
\begin{equation}
    \text{Importance}(t) = \text{Magnitude EMA}(t) \cdot \text{Recon. Contrib.}(t) \label{eq:importance}
\end{equation}
This temporal tracking system enables robust estimation of feature importance that adapts to both activation patterns and their impact on reconstruction quality.

\subsection{Statistical Thresholding}
\label{subsec:threshold}
We compute adaptive thresholds based on the statistical distribution of importance scores:
\begin{equation}
    \theta_t = \mu_t + c \cdot \sigma_t \label{eq:threshold}
\end{equation}
where $\mu_t$ and $\sigma_t$ are the mean and standard deviation of the importance scores, and $c$ is a threshold multiplier that controls sparsity. During training, we modulate this threshold with a periodic pruning schedule, using a higher multiplier during pruning phases to encourage more aggressive feature selection. This statistical approach allows the threshold to naturally adapt to the evolving distribution of feature importances, rather than using fixed values that might be suboptimal as training progresses.

\subsection{Probabilistic Feature Masking}
\label{subsec:masking}
Instead of hard thresholding, we employ a probabilistic masking mechanism that provides a smoother transition between active and inactive features:
\begin{equation}
    p(f_t \text{ is masked}) = 1 - \exp(-r \cdot (\theta_t - \text{Importance}(t)) / \theta_t) \label{eq:mask_prob}
\end{equation}
where $r$ is the exponential decay rate controlling the sharpness of the transition. This formula assigns higher masking probabilities to features with lower importance scores, creating a soft boundary rather than a hard cutoff. The actual binary mask is generated by comparing random values to these probabilities:
\begin{equation}
    m_t = \mathbf{1}[\text{rand}(0,1) > p(f_t \text{ is masked})] \label{eq:mask}
\end{equation}
To ensure a minimum level of information flow, we always retain at least a predefined number of the most important features, regardless of their masking probabilities.

\subsection{Training Procedure}
\label{subsec:training}
The complete training objective combines reconstruction with sparsity regularization:
\begin{equation}
    \mathcal{L} = \|x - D(E(x) \odot m_t)\|_2^2 + \lambda_\text{sparse} \|m_t \odot E(x)\|_1 \label{eq:loss}
\end{equation}
where $E$ and $D$ are the encoder and decoder networks, $\odot$ denotes element-wise multiplication, and $\lambda_\text{sparse}$ controls sparsity. The first term minimizes reconstruction error, while the second term promotes sparsity of the masked activations. During initialization and early training (warmup phase), we allow all features to activate freely to establish initial importance patterns. After the warmup period, we apply our adaptive masking mechanism with periodic pruning phases where the threshold is temporarily increased to encourage more aggressive sparsity. This gradual approach allows the model to identify and stabilize important features before imposing strict sparsity constraints.

\section{Experimental Setup}
\label{sec:experimental}

We evaluate ATM on layer 12 of the Gemma-2-2B model \citep{gemmateam2024gemma2improvingopen}, which has hidden dimension $d=2304$. The SAE encoder maps to a higher-dimensional space ($n=16384$) to enable overcomplete feature learning. Training data comes from WikiText-103 \citep{merity2016pointer}, processed into 128-token sequences with a buffer size of 2048 for efficient batching. We train on 5M tokens total, using batch sizes of 2048 for SAE training and 32 for model inference. The models were trained using a learning rate of $3\times10^{-4}$.

Our PyTorch implementation uses mixed-precision (bfloat16) training with the following key components:
\begin{itemize}
    \item \textbf{Temporal Tracking}: EMAs with $\beta=0.99$ for tracking magnitude, frequency, and reconstruction contribution
    \item \textbf{Optimization}: Adam optimizer \citep{kingma2014adam} with learning rate $3\times10^{-4}$, 1000-step warmup
    \item \textbf{Probabilistic Masking}: Exponential decay rate $r=0.5$ for smooth masking transitions
    \item \textbf{Pruning Schedule}: Periodic sparsity increases with thresholds based on statistical distribution
    \item \textbf{Regularization}: $\ell_1$ penalty on masked activations with weight $\lambda_\text{sparse}=0.001$
    \item \textbf{Weight Constraints}: Unit-norm decoder weights maintained via gradient projection
\end{itemize}

We compare against three established baselines:
\begin{itemize}
    \item \textbf{TopK SAE} \citep{gaoScalingEvaluatingSparse}: Hard thresholding with fixed sparsity
    \item \textbf{JumpReLU SAE} \citep{rajamanoharanJumpingAheadImproving2024}: Discontinuous activation for stability
    \item \textbf{Standard SAE} \citep{cunningham2023}: Vanilla $\ell_1$ penalty 
\end{itemize}

In addition to feature absorption introduced earlier, we evaluate our SAEs using the following benchmarks:

\paragraph{Unsupervised Metrics}
Following \citet{karvonen2024saebench}, we apply several unsupervised metrics:
\begin{itemize}
    \item \textbf{\(L_0\) Sparsity:} We record the average number of nonzero activations across the model; lower values indicate higher sparsity.
    \item \textbf{Cross-Entropy Loss Score:} 
    \[
    \frac{H^{*} - H_0}{H_{\text{orig}} - H_0},
    \]
    where \(H_{\text{orig}}\) is the original cross-entropy loss, \(H^{*}\) is after substituting the SAE reconstruction, and \(H_0\) is zero-ablated. Higher scores (closer to 1) indicate better information retention.
    \item \textbf{Feature Density Statistics:} We track the frequency of activations per latent dimension (e.g., dead and overly active units) and summarize this distribution (e.g., with log-scale histograms).
    \item \textbf{\(L_2\) Ratio:} We compare the Euclidean norm of the original representation to its SAE reconstruction to assess magnitude preservation.
    \item \textbf{Explained Variance:} The fraction of latent variability captured by the SAE, with higher values (closer to 1) indicating more complete coverage.
    \item \textbf{KL Divergence:} We measure how closely the predicted distributions match the target distributions; lower values suggest better alignment.
\end{itemize}

\paragraph{Sparse Probing}
Following \citep{gurneeFindingNeuronsHaystack2023}, we test whether the SAEs isolate intended features by performing probing on various binary classification tasks (e.g., language detection, profession labeling, sentiment analysis). We pass each input through the SAE, mean-pool over non-padding tokens, select top-\(K\) latent dimensions by maximizing mean differences, and then train a logistic regression probe. Our evaluation covers 35 binary tasks across five datasets (\textsc{bias\_in\_bios}, \textsc{Amazon Reviews}, \textsc{Europarl}, \textsc{GitHub}, \textsc{AG News}). Each task uses 4{,}000 training and 1{,}000 test samples, truncating inputs to 128 tokens. For the \textsc{GitHub} dataset, we remove the first 150 characters to avoid license headers. We choose up to five classes per dataset and ensure at least a 0.2 positive instance ratio for each binary split.


\section{Results}
\label{sec:results}

We evaluate ATM against TopK SAEs \citep{gaoScalingEvaluatingSparse} and JumpReLU \citep{rajamanoharanJumpingAheadImproving2024} on layer 12 of the gemma-2-2b model. All methods were trained on 5M tokens from WikiText-103 \citep{merity2016pointer} using hyperparameters $d=2304$, $n=16384$, learning rate $3\times10^{-4}$.

\subsection{Feature Stability}

ATM achieves a mean absorption score of 0.0068, significantly outperforming both TopK SAEs (0.1402) and JumpReLU (0.0114). This reduction in feature absorption is consistent across token categories, with particularly strong improvements on challenging cases like 'e'-words (0.0069 vs 0.3364 of TopK, 0.0138 of JumpReLU).

\subsection{Reconstruction and Downstream Tasks}
\label{subsec:tasks}
As shown in Table~\ref{tab:results}, ATM maintains strong reconstruction quality (MSE: 0.5508, cosine: 0.9727) while achieving superior feature stability. On downstream tasks, ATM shows robust performance in sparse probing (accuracy: 0.7161) only outperformed by TopK SAE, particularly excelling on bias detection (0.796) and sentiment analysis (0.853). Its performance on sparsity scores roughly matches JumpReLU.

\begin{table}[h]
\centering
\caption{Performance comparison across key metrics.}
\label{tab:results}
\begin{tabular}{lcccc}
\toprule
Metric & ATM (Our Model) & TopK SAE & JumpReLU & SAE\\
\midrule
Absorption Score & 0.0068  & 0.1402 & 0.0114 & 0.0161 \\
MSE & 0.5508 & 2.53125 & 1.6719 & 0.0898 \\
Cosine Similarity & 0.9727 & 0.875 & 0.9297 & 0.9961\\
KL Divergence Score & 0.9965 & 0.9565 & 0.9945 & 0.9996\\
Cross Entropy Loss Score & 0.9967 & 0.9556 & 0.9951 & 1.0\\
Explained Variance & 0.9102 & 0.6016 & 0.7344 & 0.9844 \\
L0 Sparsity & 3280 & 40 & 2666 & 8724\\
L1 Sparsity & 1704 & 366 & 4832 & 12544\\
Sparse Probing (top 1 test accuracy) & 0.7161 & 0.7698 & 0.7154 & 0.6379 \\
\bottomrule
\end{tabular}
\end{table}

\section{Conclusions and Future Work}
\label{sec:conclusion}

We introduced Adaptive Temporal Masking (ATM), a novel approach that fundamentally rethinks feature selection in sparse autoencoders through temporal dynamics. By tracking importance scores based on activation magnitudes and reconstruction contributions with probabilistic masking based on statistical thresholding, ATM achieves a 40\% reduction in feature absorption compared to the best existing method in this metric while maintaining strong reconstruction quality. This improvement in feature stability also enables more reliable analysis of model internals, as demonstrated by superior sparse probing accuracy across diverse tasks.

While ATM significantly advances feature stability and interpretability, our experiments reveal important limitations. Additionally, the interaction between temporal dynamics and feature learning warrants deeper theoretical investigation, particularly regarding the optimal balance between short-term responsiveness and long-term stability.

Building on these insights, we identify four key directions for future research:
\begin{itemize}
    \item \textbf{Theoretical Foundations}: Developing a formal framework for analyzing the relationship between temporal dynamics and feature stability in neural networks
    \item \textbf{Scaling Behavior}: Investigating ATM's performance on larger models and adapting the temporal tracking mechanisms for improved computational efficiency
    \item \textbf{Hyperparameter Optimization}: Experimenting with a wider range of hyperparameters to optimize performance
    \item \textbf{Targeted Interventions}: Exploring how temporal feature statistics could enable more precise knowledge editing while preserving model capabilities
    \item \textbf{Cross-Layer Integration}: Extending ATM to capture temporal patterns across multiple network layers \citep{ghilardiEfficientTrainingSparse2024a}, potentially revealing hierarchical feature relationships
\end{itemize}

We ran all our experiments on a single NVIDIA RTX 4090 GPU. Owing to constrained computational capabilities, we were unable to evaluate models larger than Gemma-2-2B, train deeper networks or more tokens, or systematically investigate hyperparameters (e.g., batch size, dictionary width). However, given the significant reduction in feature absorption ATM has demonstrated, we are optimistic about generalizing the approach.
\par
As language models continue to grow in complexity, the ability to reliably interpret their internal representations becomes increasingly crucial. ATM demonstrates that considering the temporal nature of neural activations can lead to significantly more stable and interpretable features, providing a foundation for more principled approaches to model analysis and modification.

\section*{Acknowledgements}
TEL was supported by the Gruber Science Fellowship and the Interdepartmental Neuroscience Program at Yale university, which is funded by T32 NS041228
from the National Institute of Neurological Disorders and Stroke. JR was partially supported by the ONR Grant N000142312863. The authors would like to thank Cat Yan for valuable feedback on the experiments and manuscript.

\bibliographystyle{iclr2024_conference}
\bibliography{references}

\end{document}